\documentclass[11pt]{article}

\usepackage[preprint]{acl}

\usepackage{times}
\usepackage{latexsym}

\usepackage[T1]{fontenc}

\usepackage[utf8]{inputenc}

\usepackage{microtype}

\usepackage{inconsolata}

\usepackage{graphicx}

\usepackage{booktabs}

\usepackage{xspace}
\usepackage[dvipsnames]{xcolor}
\usepackage{tabularx}

\usepackage{tcolorbox}
\tcbuselibrary{listings, skins, breakable}

\newtcblisting{rawprompt}[1][]{
  breakable,
  listing only,
  colback=gray!5,
  colframe=gray!50,
  title={#1},
  listing options={
    basicstyle=\small\ttfamily,
    breaklines=true,       
    columns=fullflexible,  
    aboveskip=0pt,
    belowskip=0pt,
    breakindent=0pt,       
    breakautoindent=false, 
  }
}

\newcommand{\bench}[0]{\textsc{FireBench}\xspace}

\title{\bench: Evaluating Instruction Following in Enterprise and API-Driven LLM Applications}

\author{
 \textbf{Yunfan Zhang\textsuperscript{1}},
 \textbf{Yijie Bei\textsuperscript{2}},
 \textbf{Jetashree Ravi\textsuperscript{2}},
 \textbf{Pawel Garbacki\textsuperscript{2}}
\\
\\
 \textsuperscript{1}Columbia University,
 \textsuperscript{2}Fireworks AI
\\
 \small{
   \textbf{Correspondence:} \href{mailto:email@domain}{yunfan.z@columbia.edu}
 }
}

\begin{document}
\maketitle

\begin{abstract}
Instruction following is critical for LLMs deployed in enterprise and API-driven settings, where strict adherence to output formats, content constraints, and procedural requirements is essential for enabling reliable LLM-assisted workflows. However, existing instruction following benchmarks predominantly evaluate natural language generation constraints that reflect the needs of chat assistants rather than enterprise users. To bridge this gap, we introduce \bench, an LLM instruction following benchmark grounded in real-world enterprise and API usage patterns. \bench evaluates six core capability dimensions across diverse applications including information extraction, customer support, and coding agents, comprising over 2,400 samples. We evaluate 11 LLMs and present key findings on their instruction following behavior in enterprise scenarios. We open-source \bench at \url{fire-bench.com} to help users assess model suitability, support model developers in diagnosing performance, and invite community contributions.
\end{abstract}

\section{Introduction}
\label{sec:introduction}
\begin{figure}[t]
    \centering
    \includegraphics[width=0.75\columnwidth]{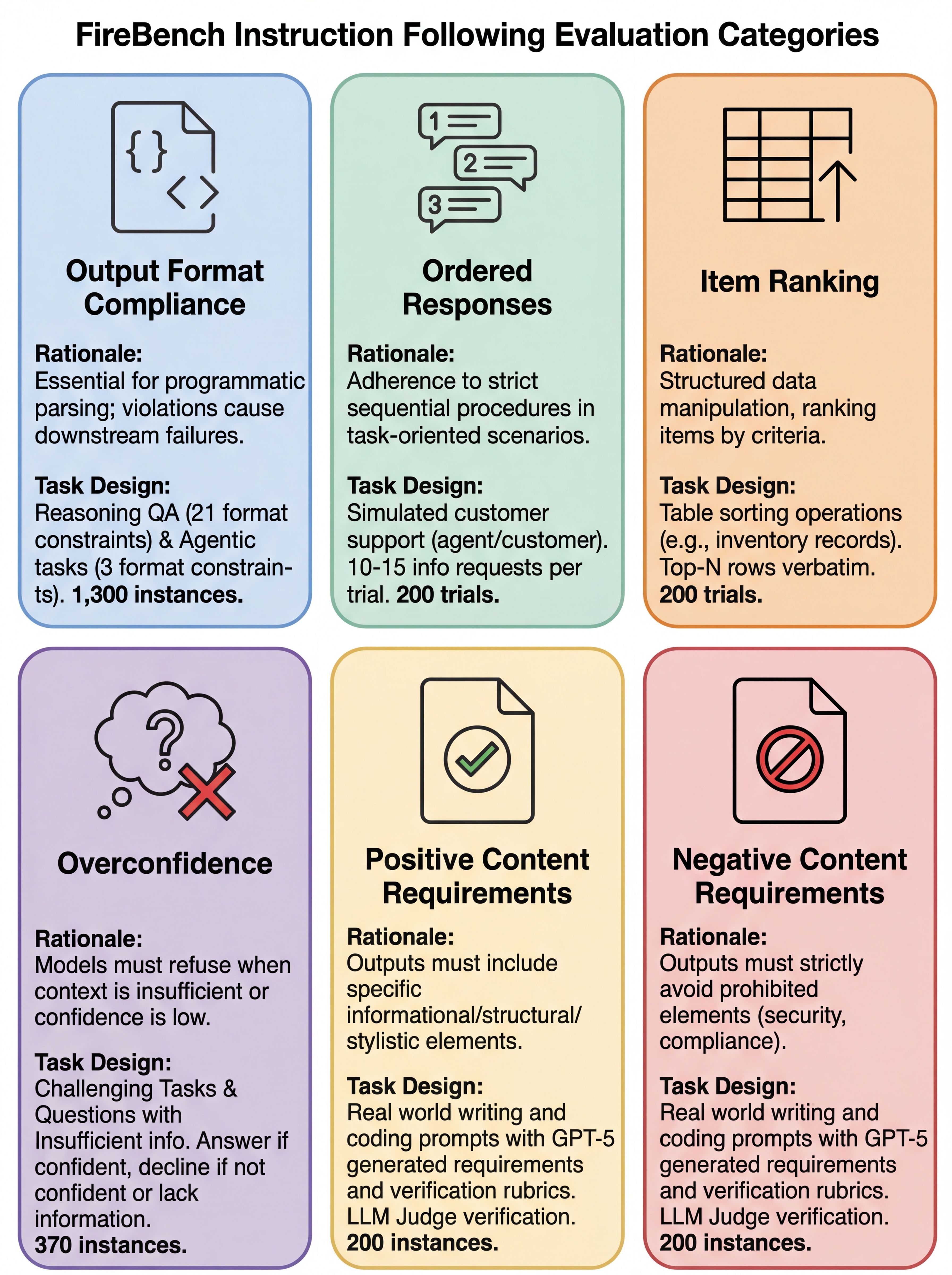}
    \caption{Overview of \bench. The benchmark evaluates six instruction-following capabilities critical to enterprise and API settings, each grounded in real-world application domains.}
    \label{fig:bench_categories}
\end{figure}
Large Language Models (LLMs) are increasingly adopted in business applications, including information extraction, decision making, code generation, and agentic workflows \citep{bommasani2021foundation,chen2021codex,dagdelen2024structured_ie,taubench,patil2024gorilla}. 
In these enterprise and API-driven settings, the ability to strictly follow instructions is paramount. For example, when an LLM is used to extract structured information from text, failing to produce outputs in a specified format can break downstream automated pipelines that rely on specific parsers \citep{patil2024gorilla}.
Similarly, in customer support or task-oriented agent scenarios, an LLM must adhere to predefined procedures step by step; deviating from required workflows can lead to task failure, even if the underlying reasoning or factual content is largely correct \citep{taubench}. 
In such environments, correctness alone is insufficient: precision, compliance, and reliability in executing the given instructions are essential.

Despite the clear need to evaluate instruction-following capabilities in enterprise and API contexts, existing benchmarks often fail to reflect these practical requirements. Many popular instruction-following benchmarks focus on natural language generation and editing constraints, such as word counts, number of paragraphs, inclusion of specific phrases, or stylistic and tonal adjustments \cite{ifeval, followbench, complexbench, followeval}. While these capabilities are useful for general-purpose chat assistants, they are largely disconnected from the needs of enterprise users, who prioritize strict adherence to response formats, deterministic ordering, constrained content, and other given requirements. As a result, current evaluations provide limited insight into whether a model can reliably operate within enterprise and API settings.

To address this gap, we introduce \textbf{\bench}, a benchmark specifically designed to evaluate instruction-following capabilities in enterprise and API-oriented use cases \cite{gpt_4p1}. The key features of our benchmark include:

\textbf{Grounded in enterprise and API usage patterns.} \bench is grounded in common usage patterns observed among enterprise and API users. We identify six core capabilities that are central to reliable instruction following in production settings: (1) \emph{output format compliance}, (2) \emph{ordered responses}, (3) \emph{item ranking}, (4) \emph{overconfidence}, (5) \emph{positive content requirements} (mandatory inclusions), and (6) \emph{negative content requirements} (mandatory exclusions). These dimensions reflect the structural and procedural constraints that real-world systems must satisfy in automated pipelines and task-oriented workflows \cite{gpt_4p1}.

\textbf{Broad coverage of enterprise applications.} Across these six dimensions, \bench covers a diverse suite of applications typical for enterprise LLM users, including reasoning with world knowledge, information extraction from long documents, customer support assistants, and coding assistants and autonomous coding agents. This diversity enables realistic assessment of instruction-following behavior under business-critical scenarios.

\textbf{Large-scale, robust evaluation.} \bench contains over 2,400 samples in total, providing substantial coverage across all capability dimensions. All categories, except positive and negative content requirements, are fully verifiable programmatically or via a simple LLM-based fuzzy matching procedure. For positive and negative requirements, we provide detailed grading rubrics to ensure consistent and transparent evaluation.

Our goal is to provide a practical tool that enables enterprise and API users to understand how well their models follow instructions in real-world enterprise settings, while also offering model developers a targeted evaluation framework to diagnose and improve instruction-following capabilities in such scenarios. We hope \bench can serve as a foundation for advancing reliable LLM deployment in enterprise production environments. We open-source \bench at \url{fire-bench.com
} and invite contributions from the research and practitioner communities.

\section{Related Work}
\label{sec:related_work}
\subsection{Instruction Following Benchmarks for LLMs}
\label{ssec:if_related_work}

A growing body of work evaluates how well LLMs follow instructions, largely in chat-assistant style natural language generation. IFEval \cite{ifeval} uses programmatically verifiable checks, but its constraints are typically surface-level (e.g., requiring certain keywords, a target language, or an approximate length). Similarly, COLLIE \cite{collie} studies generation under phrase and word-count constraints. FollowBench \cite{followbench} expands coverage with a mix of programmatic verification and GPT-4-based judging across constraint categories (e.g., content, style, and format), though many constraints remain relatively lightweight. To better reflect multi-faceted user requests, ComplexBench \cite{complexbench} emphasizes instructions with multiple constraints, and InfoBench \cite{qin-etal-2024-infobench} decomposes complex instructions into atomic yes/no rubrics scored by GPT-4. Finally, LLMBar \cite{llmbar} focuses on whether LLMs can reliably serve as evaluators for instruction-following performance.

In contrast to these benchmarks, which primarily target linguistic or stylistic constraints in free-form responses, \bench is designed around enterprise and API-driven requirements where strict output format, deterministic ordering, constrained content, and calibrated uncertainty are critical.

\subsection{LLM Benchmarks for Enterprise Settings}
\label{ssec:if_enterprise_benchmarks}

Several benchmarks evaluate LLM capabilities in enterprise and business environments. GDPVal \cite{patwardhan2026gdpval} and OfficeBench \cite{wang2024officebench} assess whether models can complete office-related tasks within productivity software such as Microsoft Office. CRMArena \cite{huang-etal-2025-crmarena} focuses on customer service scenarios, evaluating LLMs as support agents. Complementing these efforts, \bench specifically targets instruction-following robustness in enterprise settings.

\section{Methodology}
\label{sec:method}

\begin{table}[t]
\centering
\footnotesize
\setlength{\tabcolsep}{3pt}
\begin{tabularx}{\columnwidth}{p{1.4cm} X r}
\toprule
\textbf{Category} & \textbf{Task Design} & \textbf{\#} \\
\midrule
Output Format (\S\ref{ssec:output_format})
  & Reasoning and long-document QA under 21 format constraints; coding problems under 3 structured formats. Verified programmatically.
  & 1{,}300 \\
\addlinespace
Ordered Responses (\S\ref{ssec:ordered_responses})
  & Simulated multi-turn customer support sessions in which the agent must collect 10--15 information fields in strict sequential order. Results are verified programmatically.
  & 200 \\
\addlinespace
Item Ranking (\S\ref{ssec:item_ranking})
  & Table sorting operations over realistic records; models must return top-$N$ rows verbatim according to specified sorting criteria. Results are verified programmatically.
  & 200 \\
\addlinespace
Over\-confidence (\S\ref{ssec:overconfidence})
  &  \emph{Challenging tasks}: paired normal and uncertainty-aware prompts from challenging reasoning benchmarks. \emph{Insufficient information}: unanswerable questions over long-form documents requiring explicit refusal. Models are evaluated on whether they appropriately decline questions they cannot solve or for which sufficient information is unavailable.
  & 370 \\
\addlinespace
Positive Content (\S\ref{ssec:positive_content_requirements})
  & Selected Arena Hard 2.0 prompts augmented with mandatory content requirements. Evaluated via LLM judge.
  & 200 \\
\addlinespace
Negative Content (\S\ref{ssec:negative_content_requirements})
  & Same prompts augmented with prohibitive constraints; models must avoid forbidden elements. Evaluated via LLM judge.
  & 200 \\
\midrule
\multicolumn{2}{l}{\textbf{Total}} & \textbf{2{,}470} \\
\bottomrule
\end{tabularx}
\caption{Composition of \bench{}. Each category targets a distinct instruction-following capability critical to production deployments.}
\label{tab:benchmark_composition}
\end{table}

We design \bench~around six core instruction-following capabilities
critical to production deployments: (1)~\emph{output format compliance},
(2)~\emph{ordered responses}, (3)~\emph{item ranking},
(4)~\emph{overconfidence}, (5)~\emph{positive content requirements},
and (6)~\emph{negative content requirements}. These categories capture
common practical constraints that real-world systems must
satisfy within automated pipelines and task-oriented
workflows~\cite{gpt_4p1}. An overview of the benchmark composition is
provided in Table~\ref{tab:benchmark_composition}. We additionally
include a sample input for each evaluation category in
Appendix~\ref{sec:bench_examples_appendix}.

We emphasize that \bench~is designed to be \emph{representative and
realistic, not exhaustive}. Many enterprise instruction-following
requirements remain uncovered; we welcome future community
contributions to expand the benchmark's scope.

In the remainder of this section, we detail the motivating rationale and task design for each of the six evaluation categories.

\subsection{Output Format Compliance}
\label{ssec:output_format}

\subsubsection{Rationale}
Format adherence is essential for information extraction, agentic
systems, and reinforcement learning from verifiable rewards (RLVR).
These applications require programmatic parsing of LLM outputs; format
violations cause downstream failures even when the underlying answers
are factually correct. Given the prevalence of such systems in
enterprise settings, output format compliance constitutes one of our
benchmark's most critical categories.

\subsubsection{Task Design}
We model enterprise usage through two paradigms: \emph{Question
Answering} (QA) and \emph{Agentic Interactions}.

For \emph{QA tasks}, we focus on information extraction and
reasoning--the most common enterprise use cases. We sample 25 questions
each from long-document understanding benchmarks
(LongBench~V2~\cite{longbench} and QUALITY~\cite{quality}) and
reasoning benchmarks (GPQA Diamond~\cite{gpqa} and
LogiQA~\cite{logiqa}). Each sample is evaluated under 21 distinct
format constraints, including standard markup languages (JSON, XML),
common delimiters (e.g., \texttt{\textbackslash boxed\{\}}), and
adversarial variants (e.g., \texttt{\textbackslash boxed[ ]}),
yielding a total of 1{,}000 evaluation instances. All output formats
are verified programmatically to minimize evaluation errors.

For \emph{Agentic Interaction tasks}, we use coding-agent prompts as a
proxy for agentic workflows. We sample 100 problems from the MHPP
dataset~\cite{dai2024mhpp} as the input set. Models must separate
\emph{reasoning}, \emph{code}, and \emph{explanation} under one of
three constrained output formats (JSON, XML, or Markdown), yielding
300 instances. This design reflects practical requirements in
LLM-assisted software development, where modular, machine-readable
outputs are necessary for both execution and inspection.

\subsection{Ordered Responses}
\label{ssec:ordered_responses}

\subsubsection{Rationale}
In task-oriented scenarios such as customer support, LLMs must adhere
to predefined procedures in a strict sequential order. Deviations from
the prescribed workflow can cause failures even when the underlying
reasoning is correct. This category evaluates a model's ability to
follow step-by-step instructions faithfully.

\subsubsection{Task Design}
We evaluate ordered-response capabilities by simulating customer
support sessions in which the same model assumes dual roles in separate contexts: a support
agent and a customer. The agent receives an ordered list of 10--15
information requests and must collect them strictly in sequence, posing
exactly one question per turn. The customer responds with the
corresponding ground-truth value for each request, providing no
additional information.

We conduct 200 trials for this category. 
Each trial samples 10--15 fields from a pool of approximately 70
fields, and one ground truth answer per field from 5--10 human-vetted answers. We then evaluate whether models maintain ordering constraints,
execute multi-turn interactions correctly, and return the requested
information without modification or information leakage. Results are
verified against the ground truth programmatically.

\subsection{Item Ranking}
\label{ssec:item_ranking}

\subsubsection{Rationale}
Structured data manipulation is ubiquitous in data-centric
applications. Models are frequently required to rank items according
to specified criteria, making this a key instruction-following
capability for enterprise use.

\subsubsection{Task Design}
We evaluate item ranking by simulating table-sorting operations.
Models receive realistic tables (e.g., inventory records, product
listings, financial summaries) together with a designated sorting
attribute and must return the top-$N$ rows verbatim according to the
specified criterion—equivalent to
\texttt{ORDER BY <attribute> DESC LIMIT N} in SQL. We construct 20
distinct tables with randomly assigned sorting attributes and evaluate
models across 200 trials to ensure robustness. Evaluation is again
conducted programmatically.

\subsection{Overconfidence}
\label{ssec:overconfidence}

\subsubsection{Rationale}
Overconfidence represents a critical failure mode in high-stakes
applications. Enterprise developers often instruct models to refuse when the available context is insufficient or confidence is
low, thereby enabling escalation to human supervisors. This category
evaluates whether models can appropriately abstain from answering
under uncertainty.

\subsubsection{Task Design}
We define two subcategories: \emph{challenging tasks} and
\emph{insufficient information}.

For \emph{challenging tasks}, we randomly sample a total of 300
questions from GPQA Diamond~\cite{gpqa}, Humanity's Last
Exam~\cite{hle}, and SimpleQA~\cite{simpleqa}. Each question is
subjected to two independent inferences: (a)~a standard prompt
requesting an answer, and (b)~an uncertainty-aware prompt instructing
the model to answer only if confident and to explicitly decline
otherwise. A response is scored as correct if and only if (i)~the
model answers correctly under the standard prompt and does not decline
under the uncertainty-aware prompt, or (ii)~the model answers
incorrectly under the standard prompt but explicitly declines under
the uncertainty-aware prompt. This design measures whether models can
precisely decline the questions they are unable to solve, without
degrading their performance on easier or well-informed tasks.

For \emph{insufficient information}, we manually curate
${\sim}$70 samples, each comprising a long-form document, a topically
related but unanswerable question, and an instruction to answer only
when sufficient evidence is present in the document. A correct
response requires an explicit refusal; any attempted answer
constitutes failure.

\subsection{Positive Content Requirements}
\label{ssec:positive_content_requirements}

\subsubsection{Rationale}
Enterprise applications often require outputs to include specific
informational, structural, or stylistic elements to ensure alignment
with business requirements, regulatory compliance, or downstream
processing expectations. Missing even a single mandatory element can
render an otherwise correct output unusable.

\subsubsection{Task Design}
We evaluate positive content adherence using challenging prompts drawn
from Arena Hard Auto~2.0~\cite{arenahard}. We manually select 200
prompts spanning programming and writing tasks, then employ GPT-5 to
generate supplementary content requirements along with corresponding
verification rubrics. These requirements specify mandatory output
elements (e.g., particular code patterns, required informational
content). Model outputs are evaluated for compliance using GPT-4.1 as
an LLM judge. A violation of any individual rubric is treated as task
failure.

\subsection{Negative Content Requirements}
\label{ssec:negative_content_requirements}

\subsubsection{Rationale}
Complementary to positive requirements, enterprise applications
frequently impose strict prohibitions on model outputs. Such negative
constraints arise from security policies, regulatory frameworks,
quality standards, and business logic. Violating these constraints can
introduce security vulnerabilities, compliance risks, or system
incompatibilities, making the ability to respect prohibitions critical
for production deployment.

\subsubsection{Task Design}
Using the same 200 Arena Hard Auto~2.0 prompts, we augment each with
negative constraints and verification rubrics generated by GPT-5.
These constraints prohibit specific operations, formats, stylistic
elements, or content types. Models must fulfill the original task
objectives while strictly avoiding all forbidden elements. Outputs are
evaluated for compliance using GPT-4.1 as an LLM judge; any violation
constitutes failure.

\section{Results}
\label{sec:results}

\begin{table*}[t]
\centering
\footnotesize
\renewcommand{\arraystretch}{1.15}
\setlength{\tabcolsep}{4pt} 
\begin{tabular}{@{}lccccccc@{}}
\toprule
\textbf{Model} &
\textbf{Format} &
\textbf{\shortstack{Ordered\\Responses}} &
\textbf{Ranking} &
\textbf{Overconfidence} &
\textbf{\shortstack{Positive\\Content\\Requirements}} &
\textbf{\shortstack{Negative\\Content\\Requirements}} &
\textbf{Overall} \\
\midrule
DeepSeek V3.1 Terminus & 54.3 & 65.5 & \underline{92.5} & \textbf{86.0} & 84.0 & 81.4 & \textbf{74.0} \\
GPT-5.1 Medium Thinking & 62.2 & 53.5 & \textbf{93.0} & \underline{67.7} & 86.5 & 84.0 & \underline{72.7} \\
GPT-4.1 & \textbf{86.9} & \textbf{76.3} & 32.5 & 38.6 & \textbf{94.5} & \textbf{94.5} & 70.5 \\
Qwen3 235B Thinking 2507 & 39.9 & \underline{72.0} & 76.0 & 42.0 & 85.5 & 84.2 & 66.6 \\
Qwen3 235B Instruct 2507 & \underline{65.6} & 69.8 & 38.0 & 44.2 & 87.0 & \underline{93.5} & 66.3 \\
Claude Sonnet 4.5 & 64.0 & 59.5 & 63.0 & 53.4 & 79.0 & 74.5 & 65.3 \\
gpt-oss-120b & 55.3 & 47.8 & 71.0 & 42.9 & \underline{91.5} & 78.5 & 64.5 \\
Kimi K2 Thinking & 54.6 & 61.0 & 72.5 & 32.8 & 81.0 & 86.0 & 63.2 \\
Kimi K2 Instruct 0905 & 56.6 & 63.0 & 8.5 & 65.6 & 90.0 & \textbf{94.5} & 62.1 \\
Llama 4 Maverick Instruct & 57.9 & 61.5 & 27.9 & 35.6 & 80.5 & 91.5 & 59.1 \\
GPT-5.1 Instant & 63.3 & 52.5 & 16.0 & 62.4 & 70.5 & 82.5 & 58.6 \\
\bottomrule
\end{tabular}
\caption{Evaluation results on \bench. Numbers denote the percentage of samples in which the model correctly performed the expected action in each category, as well as the overall score, computed as the average across categories. Higher is better. For each category and the overall score, the best result is shown in \textbf{bold}, and the second best is \underline{underlined}. Results indicate that instruction following in enterprise and API settings remains challenging, with no model achieving more than 75\% overall.}
\label{tab:fireworks-instruction-following}
\end{table*}

\begin{figure}[t]
    \centering
    \includegraphics[width=0.78\columnwidth]{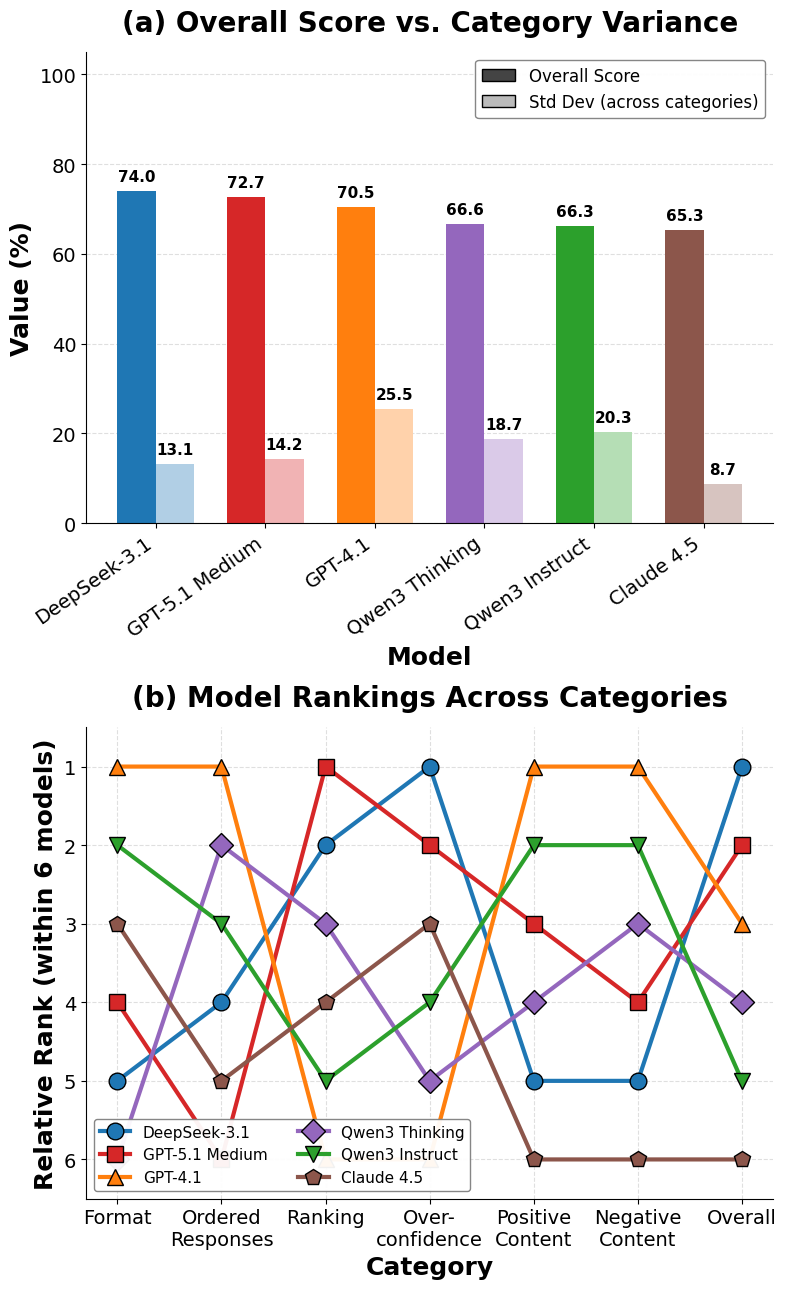}
    \caption{(a) Overall score and standard deviation across categories for the top-6 models by overall performance. All models exhibit high variance, indicating inconsistent instruction-following ability across categories. (b) Relative ranking of the same six models within each category. Model rankings shift substantially from one category to another, demonstrating that no single model dominates uniformly.}
    \label{fig:variance-and-rankings}
\end{figure}

As shown in Table~\ref{tab:fireworks-instruction-following}, we evaluate 11 closed and open-weight frontier models on \bench{} and report their per-category and overall performance. We summarize our key findings below.

\paragraph{Precise instruction following for enterprise and API settings remains a hard challenge.}
Despite the rapid progress in LLMs capabilities, our results suggest that reliable instruction following in enterprise and API contexts is far from solved. The best-performing model, DeepSeek-V3.1, achieves only 74.0\% overall, while the second-best, GPT-5.1 Medium Thinking, reaches 72.7\%. No model surpasses the 75\% threshold, and the majority of evaluated models score below 67\%. These results underscore a significant gap between the perceived capabilities of frontier models and their actual compliance with precise instructions--the very type of instructions that are critical for reliable deployment in production API pipelines.

\paragraph{Large variance across instruction-following categories.}
As illustrated in Figure~\ref{fig:variance-and-rankings}, we visualize the overall scores alongside the cross-category standard deviation for the top-6 models, as well as their relative rankings within each category. The variance across categories is striking: even the top-ranked DeepSeek V3.1 exhibits a standard deviation of 13.1 percentage points, while GPT-4.1 shows the highest variance at 25.5 points--reflecting its dominant performance on Format (86.9\%) and Content Requirements ($\geq$94.5\%) but sharp drops on Ranking (32.5\%) and Overconfidence (38.6\%). The ranking plot further highlights this instability: model rankings shift dramatically across categories. For instance, GPT-4.1 ranks 1st on Format, Ordered Responses, and both Content Requirements categories, yet drops to 5th or 6th on Ranking and Overconfidence. Conversely, DeepSeek V3.1 ranks 1st on Overconfidence but only 4th on Format. These findings suggest that selecting a model for a particular enterprise deployment requires careful consideration of which instruction-following capabilities are most critical for the target use case, as overall scores alone can mask substantial category-level weaknesses.

\paragraph{Reasoning models generally outperform their non-reasoning variants.}
Our benchmark includes three pairs of reasoning and non-reasoning model variants, enabling a controlled comparison. In all three cases, the reasoning variant achieves a higher overall score: GPT-5.1 Medium Thinking (72.7\%) outperforms GPT-5.1 Instant (58.6\%) by a margin of 14.1 points; Kimi K2 Thinking (63.2\%) edges out Kimi K2 Instruct (62.1\%); and Qwen3 235B Thinking (66.6\%) narrowly surpasses Qwen3 235B Instruct (66.3\%). A closer inspection reveals that the largest gains from reasoning consistently appear in the \emph{Ranking} category, where we frame evaluation items as a table ranking problem requiring models to sort and order structured outputs. Here, GPT-5.1 Medium Thinking scores 93.0\% compared to only 16.0\% for GPT-5.1 Instant, an improvement of 77.0 percentage points. Similarly, Qwen3 Thinking doubles the score of its Instruct counterpart (76.0\% vs.\ 38.0\%), and Kimi K2 Thinking achieves 72.5\% versus a mere 8.5\% for the Instruct variant. This pattern suggests that non-reasoning models, which lack an explicit chain-of-thought process, struggle to correctly sort and rank structured outputs, a task that inherently requires multi-step comparison and ordering.

\paragraph{Why models still struggle with formatting.}
Perhaps surprisingly, formatting, ostensibly one of the more mechanical aspects of instruction following, remains a persistent challenge. Even the best-performing model on this category, GPT-4.1, achieves only 86.9\%, while several strong models score below 60\%. We hypothesize that this brittleness stems from the narrow distribution of formatting conventions encountered during post-training. Models are extensively trained to produce outputs in common formats such as \texttt{\textbackslash boxed\{\}}, and indeed perform near-perfectly when these familiar formats are requested. However, \bench{} also includes adversarial formatting variants (e.g., \texttt{\textbackslash boxed[ ]}) that are syntactically similar but deviate from training conventions. An examination of our experiment logs confirms this hypothesis: GPT-4.1 and Qwen3 235B Instruct, the two highest-scoring models on formatting, both achieve 100\% accuracy on the standard \texttt{\textbackslash boxed\{\}} format, but their performance drops sharply to only 53\% and 73\%, respectively, on the adversarial variant. This gap reveals that current models tend to memorize specific formatting patterns from post-training rather than developing a generalizable ability to follow arbitrary formatting instructions.

\section{Conclusions}
We introduced \bench, a practical instruction-following benchmark grounded in real-world enterprise and API usage patterns. Unlike existing benchmarks that primarily target natural language generation constraints for chat assistants, \bench evaluates six core instruction following capability dimensions under diverse enterprise applications. Our evaluation of 11 LLMs reveals that precise instruction following in enterprise settings remains a significant challenge, and performance varies dramatically across categories.

\clearpage
\newpage

\section*{Limitations}
Although \bench aims to evaluate LLM instruction following capabilities under a set of representative instruction following categories and applications for enterprise settings, our coverage is far from covering the complex needs that occur in real-world applications. Looking ahead, we plan to expand \bench with additional categories and application scenarios, potentially through community contributions. We are also interested in exploring composable constraints \cite{complexbench}, where multiple requirements from different categories must be satisfied simultaneously, better reflecting the complexity of real-world enterprise LLM applications.


\bibliography{custom}

\appendix

\section{\bench Examples}
\label{sec:bench_examples_appendix}

\subsection{Output Format Compliance Examples}
\begin{rawprompt}
Please solve the problem presented below by thinking step by step. Put your final answer at the end of your response. Your final answer should be a single upper-case letter in \\boxed[] (e.g. \\boxed[A]).

A chemist performed two reactions by taking two unknown compounds and treated them separately with two different reducing agents. Select the proper starting material for both of the reactions.

A + LiBH4 + H+ ---> (R)-4-ethyltetrahydro-2H-pyran-2-one B + BH3 + H+ ---> (S)-4-ethyltetrahydro-2H-pyran-2-one

(A) A = (S)-3-ethyl-5-isobutoxy-5-oxopentanoic acid , B = (S)-3-ethyl-5-isobutoxy-5-oxopentanoic acid

(B) A = (R)-3-ethyl-5-isobutoxy-5-oxopentanoic acid, B = (S)-3-ethyl-5-isobutoxy-5-oxopentanoic acid

(C) A = (R)-3-ethyl-5-isobutoxy-5-oxopentanoic acid , B = (R)-3-ethyl-5-isobutoxy-5-oxopentanoic acid

(D) A = (S)-3-ethyl-5-isobutoxy-5-oxopentanoic acid, B = (R)-3-ethyl-5-isobutoxy-5-oxopentanoic acid
\end{rawprompt}

\subsection{Ordered Responses Examples}
\begin{rawprompt}
Please collect the following list of information in EXACTLY this order, one at a time:

Full Name
Email Address
Address
Years at the company
Date of Birth

Instructions:
- You should formulate the questions in a natural manner, using your own words.
- Ask only ONE question at a time. Wait for the user's response before asking the next question.
- Ask the questions in the exact order listed above.
- Do not ask any additional questions.
- Once you have asked all questions and received all answers, say "Thank you for providing all the information."

Start by asking the first question now.
\end{rawprompt}

\subsection{Item Ranking Examples}
\begin{rawprompt}
Please sort the following table by the column '{first_selected_column}', in a {order} order. {return_rows_prompt}.

You must follow the following instructions carefully.
- You must exactly recite the entire selected rows in the table, in a CSV format, not including the headers.

- When sorting the table by columns that are not entirely numeric, you should sort the tables using the dictionary order.\
You should still respect the "descending" or "ascending" order specified in the first line of your instructions. The rules for dictionary order are as follows:
1. Compare characters left-to-right.
2. At the first position where they differ, the string with the smaller character comes first.
3. If all compared characters are equal but one string ends, the shorter (the prefix) comes first.
4. For character ordering, use ASCII code point order: '0' < '1' < ... < '9' < 'A' < ... < 'Z' < 'a' < ... < 'z'

- You should approach this problem step by step, and put your final results after "Final Result:". Use the following format:
Final Result:
ENR1234,Fall 2023,CS101,Intro to Computer Science,Computer Science,3,150,Dr. John Doe
ENR5678,Fall 2023,MATH201,Calculus I,Mathematics,4,120,Prof. Jane Smith
... Any additional rows that are selected ...

The table is as follows:
TransactionID,Region,ProductCategory,ProductName,UnitsSold,UnitPrice,DiscountRate,TotalRevenue
TX9F2,North,Electronics,Smartphone A1,125,399.99,5.2,47400
R7KQ1,East,Home Appliances,Vacuum Cleaner V7,68,149.50,7.5,9420
M2D8X,West,Clothing,Winter Jacket W3,210,89.90,10.0,17000
Q8L5C,South,Electronics,Laptop L5,45,799.00,3.8,34600
N5B7Z,North,Groceries,Olive Oil 1L,330,7.49,2.5,2410
P1X4R,East,Furniture,Office Chair C2,57,129.00,8.2,6740
W3J9S,West,Clothing,Sneakers S9,195,59.95,5.0,11100
S6F2A,South,Home Appliances,Microwave M3,74,199.00,6.7,13700
Y2L8Q,North,Electronics,Tablet T8,92,329.00,4.4,28900
E4K7M,East,Groceries,Coffee Beans 1kg,285,14.50,3.2,4000
H3T5V,West,Furniture,Dining Table D4,33,499.00,9.0,14900
K8R2L,South,Clothing,T-Shirt Basic B1,410,19.99,7.5,7570
D5M7C,North,Electronics,Headphones H2,160,79.90,6.0,12000
A2J9P,East,Home Appliances,Blender BL1,95,89.00,4.8,8050
V6K3X,West,Groceries,Pasta Pack 500g,420,2.79,2.0,1140
Z1N4F,South,Furniture,Bed Frame BF7,28,699.00,8.5,17900
C9Y6B,North,Clothing,Jeans Classic J3,190,49.95,5.1,8990
G7U3W,East,Electronics,Smartwatch SW4,105,229.00,7.0,22400
R4Q8T,West,Groceries,Rice Bag 5kg,138,11.50,3.3,1530
L2F9M,South,Home Appliances,Air Conditioner AC9,21,1199.00,10.0,22700
J5C1H,North,Furniture,Bookshelf B2,42,219.00,5.6,8690
T7K6S,East,Clothing,Summer Dress SD6,175,39.90,6.8,6520
B8M4N,West,Electronics,Camera C7,54,549.00,4.5,28300
X3P7Q,South,Groceries,Milk Carton 1L,495,1.39,2.2,673
M1D6A,North,Clothing,Socks Pair SP1,520,4.99,5.5,2460
U9Z3E,East,Furniture,Office Desk OD2,31,429.00,9.3,12000
Q4L7K,West,Home Appliances,Toaster T6,84,59.50,4.1,4800
S8V1F,South,Electronics,Monitor 24M,63,179.00,5.0,10700
F2H9W,North,Groceries,Bottle Water 6pk,280,3.49,2.7,952
Y6C3J,East,Clothing,Jacket Light JL2,142,59.90,7.4,7870
D1X8R,West,Electronics,Game Console G9,29,449.00,3.5,12500
K9M2B,South,Furniture,Couch Set CS5,19,899.00,9.9,15400
N7Z4P,North,Home Appliances,Refrigerator RF8,14,1499.00,8.0,19300
V3J6L,East,Groceries,Cheese Block 500g,192,6.49,4.6,1190
R8K1T,West,Clothing,Formal Shirt FS3,155,34.90,5.2,5130
P6H9C,South,Electronics,Drone D1,37,699.00,6.3,24300
A5U7Y,North,Furniture,Lamp Stand L2,66,89.00,3.8,5650
J2F4X,East,Clothing,Hat H5,210,14.99,8.1,2890
L9Q3S,West,Groceries,Orange Juice 1L,305,2.99,2.9,885
G1Z5M,South,Home Appliances,Washing Machine WM4,23,799.00,7.7,17000
\end{rawprompt}

\subsection{Overconfidence Examples}
\begin{rawprompt}
Please solve the problem presented below by thinking step by step. Put your final answer at the very end of your response. Your final answer should be a single upper-case letter after Answer: (e.g. Answer: A). However, if you are unable to solve the problem or not sure about your answer, reply exactly with \"Answer: Skip\".

A chemist performed two reactions by taking two unknown compounds and treated them separately with two different reducing agents. Select the proper starting material for both of the reactions.

A + LiBH4 + H+ ---> (R)-4-ethyltetrahydro-2H-pyran-2-one B + BH3 + H+ ---> (S)-4-ethyltetrahydro-2H-pyran-2-one

(A) A = (S)-3-ethyl-5-isobutoxy-5-oxopentanoic acid , B = (S)-3-ethyl-5-isobutoxy-5-oxopentanoic acid

(B) A = (R)-3-ethyl-5-isobutoxy-5-oxopentanoic acid, B = (S)-3-ethyl-5-isobutoxy-5-oxopentanoic acid

(C) A = (R)-3-ethyl-5-isobutoxy-5-oxopentanoic acid , B = (R)-3-ethyl-5-isobutoxy-5-oxopentanoic acid

(D) A = (S)-3-ethyl-5-isobutoxy-5-oxopentanoic acid, B = (R)-3-ethyl-5-isobutoxy-5-oxopentanoic acid
\end{rawprompt}

\subsection{Positive Content Requirement Examples}
\begin{rawprompt}
Given a postgres table with id, name, parent_id and rank.  Assume leaves can have variable number of ancestors each with variable ranks.  Select all rows with one column per observed rank, filling in name, or nil, when the corresponding ancestor exists. Include exactly one complete SQL query labeled \"Final SQL:\". Provide a single sentence explaining how ranks are pivoted into columns, explicitly stating that nil is represented as SQL NULL, and list the exact output column names used by the query.

Scoring criteria (not given to the tested model): - The response contains exactly one SQL query and it is labeled \"Final SQL:\".\n- The response includes exactly one sentence explaining how ranks are pivoted into columns.\n- The response explicitly states that nil is represented as SQL NULL.\n- The response lists the exact output column names used by the query.
\end{rawprompt}

\subsection{Negative Content Requirement Examples}
\begin{rawprompt}
Given a postgres table with id, name, parent_id and rank.  Assume leaves can have variable number of ancestors each with variable ranks.  Select all rows with one column per observed rank, filling in name, or nil, when the corresponding ancestor exists. Do not include any explanations, step-by-step reasoning, sample data, or result tables; output only a single PostgreSQL SQL SELECT query. Do not use non-core extensions (e.g., ltree or tablefunc/connectby), and do not create temporary tables, views, functions, or procedures. Do not wrap the query in code fences or include SQL comments.

Scoring criteria (not given to the tested model): - The response contains exactly one SQL SELECT statement and no surrounding prose or examples.\n- The response does not include CREATE/INSERT/ALTER/DROP statements, or any temp tables, views, functions, or procedures.\n- The response does not reference extensions or modules such as ltree or tablefunc/connectby.\n- The output is not wrapped in Markdown code fences and contains no SQL comments.
\end{rawprompt}

\end{document}